\documentclass{article}
\usepackage{amsmath}
\usepackage{amsthm}
\usepackage{amssymb}
\usepackage[a4paper,vscale=0.8,hscale=.75]{geometry}
\usepackage[round]{natbib}
\usepackage[colorlinks=true,allcolors=blue!20!black]{hyperref}

\date{\today\quad (preprint)}

\newenvironment{frontmatter}{}{}
\newenvironment{keyword}{}{}
\newcommand{\kwd}[1]{#1}

\newcommand{\presentauthors}{\author{Tomoyuki Kaneko\thanks{kaneko@acm.org} and Shuhei Yamashita\\
  Graduate School of Arts and Sciences, the University of Tokyo, Japan}\maketitle}
\newcommand{\configurebibstyle}{\bibliographystyle{unsrtnat}}
 
\newcommand{\gamevariant}[2]{\textit{2048}$_{#1\times #2}$}
\newcommand{\thisgame}{\gamevariant{4}{3}}
  
\usepackage{tikz}
\usetikzlibrary{automata, arrows.meta}
\usepackage{subcaption}
\usepackage{booktabs}

\usepackage{graphicx}
\graphicspath{{..}}

\theoremstyle{definition}
\newtheorem{definition}{Definition}[section]

\begin{document}

\begin{frontmatter}
  \title{Strongly Solving 2048${}_{4\times3}$}
  \presentauthors{}

\begin{abstract}
\textit{2048} is a stochastic single-player game involving 16 cells on a 4 by 4 grid, where a player chooses a direction among up, down, left, and right to obtain a score by merging two tiles with the same number located in neighboring cells along the chosen direction. 
This paper presents that a variant \textit{\thisgame} with 12 cells on a 4 by 3 board, one row smaller than the original, has been strongly solved.  
In this variant, the expected score achieved by an optimal strategy is about $50724.26$ for the most common initial states: ones with two tiles of number 2.  
The numbers of reachable states and \textit{afterstates} are identified to be $1,152,817,492,752$ and $739,648,886,170$, respectively. 
The key technique is to partition state space by the sum of tile numbers on a board, which we call the \textit{age} of a state.  
An age is invariant between a state and its successive afterstate after any valid action and is increased two or four by stochastic response from the environment.  Therefore, we can partition state space by ages and
enumerate all (after)states of an age depending only on states with the recent ages.  
Similarly, we can identify (after)state values by going along with ages in decreasing order. 
\end{abstract}


\end{frontmatter}

\section{Introduction}

\textit{2048} is a popular video game
and also a challenging environment for reinforcement learning agents as investigated by, e.g., \cite{SzubertJaskowski2014, antonoglou2022planning}.
While the original game uses a $4\times 4$ board, this paper discusses a variant with a $4\times 3$ board.
Let \gamevariant{r}{c} be a variant with rows $r$ and/or columns $c$.  
In 2022, \gamevariant{3}{3} (or Mini2048) was strongly solved by~\cite{shuhei22:_stron_solvin_board_perfor_evaluat}.  
Since then, it has served as a tractable evaluation
environment for reinforcement learning agents, for example, by \cite{matsuzaki10645595}.  

\begin{figure}[t]
  \includegraphics[width=\linewidth]{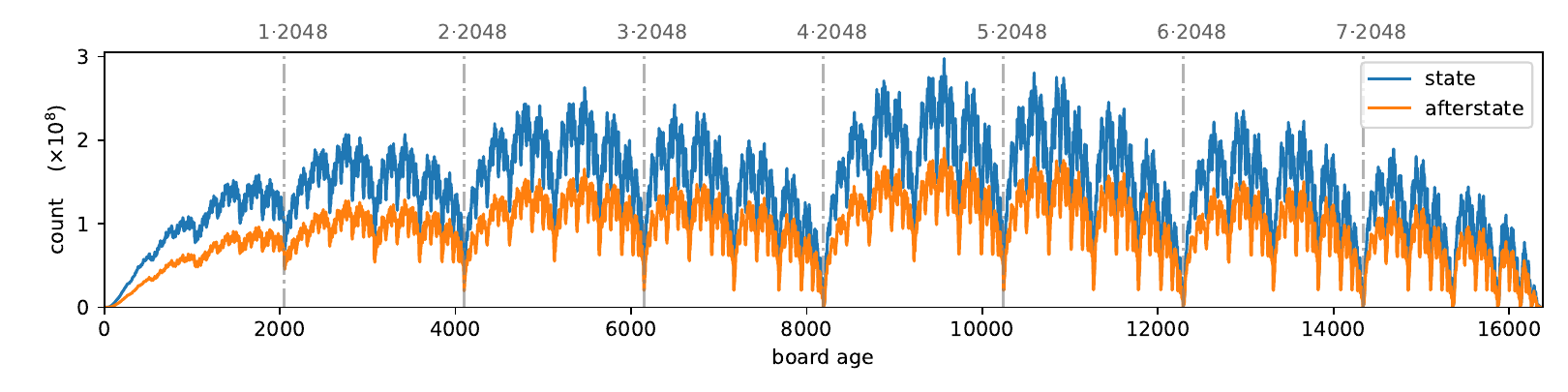}
  \caption{Number of states and afterstates per age.  The number of states is at most $297,784,736 < 3\cdot 10^6$ for each age, so we can keep several ages in memory.   Multiple valleys were observed at ages near multiple of 2048, shown by vertical dashed lines.  Apparently, they indicate the difficulty of making tile 2048, which requires careful arrangement of 10 tiles ($2, 4, 8, 16, 32, 64, 128, 256, 512,$ and $1024$) while there are only 12 squares.}
  \label{fig:counts}
\end{figure}

This paper presents that a larger variant \gamevariant{4}{3} was strongly solved. 
The results tell us that 
the numbers of states and \textit{afterstates} are $1,152,817,492,752$ and $739,648,886,170$, respectively, in this variant. 
The numbers are more than $10^4$ times larger than those of \gamevariant{3}{3} where the number of states is merely $48,713,519$.
The number of states in \thisgame{} is huge enough to prohibit straightforward approaches from enumerating them all.  
The key idea that enabled the computation, in a few days on a personal computer is to partition state space by the sum of the numbers on a board, which we call the \textit{age} of a state.  
An age is invariant between a state and its successive afterstate after any valid action and is increased two or four by stochastic response from the environment.  Therefore, we can partition state space by ages and
enumerate all (after)states of an age depending only on states with the recent ages.  
As shown in Fig.~\ref{fig:counts}, the number of (after)states is found to be less than $3\cdot 10^6$ for each age.  Therefore, it is feasible to keep several consecutive ages in memory assuming gigabytes of memory that are now commonly available. 
Similarly, we can identify (after)state values by going along with ages in decreasing order. 
In addition to the partitioning, we introduce a compact representation adapting Elias-Fano codes to reduce disk usage. 
While a straightforward store requires $739,648,886,170 \cdot 48 / 8 \approx 4.4$ TiB, our representation did it in about $1.4$ TiB in a search-efficient manner and in less than $300$~GiB if specialized for optimal playing.  

We also present several statistics on optimal value function identified verifying game players intuitions. 
Note that this paper is an extended version from a workshop proceeding written by the same authors published in Japanese.\footnote{\url{http://id.nii.ac.jp/1001/00229224/}}  Although the primary idea of age and basic statistics were presented in the earlier version, our method have been substantially improved including the compact state representation and the overall efficiency.  Moreover, several experiments help understanding of the game have been augmented.  

\section{Backgrounds and related work}
\subsection{Markov decision process and 2048}

We briefly introduce the stochastic Markov decision process (MDP), and then define game of \textit{2048} following the notation.  
An initial state $s_0$ is given among a set of initial states $\mathcal{S}_0$.  
For each time step $t \ge 0$, an agent chooses an action $a_t$ from a set of valid actions $A(s_t)$ in the current state $s_t$.
The state at the next time step $s_{t+1}$ and reward $r_{t+1}$ are sampled following the transition probability conditioned on $s_{t}$ and $a_t$. 

In the original \textit{2048}, a state consists of 16 cells in a $4\times 4$ square grid.  In our variant \thisgame, a state consists of 12 cells.  
Each cell has a tile with number of power of two, i.e., $2, 4, 8, \cdots, 2^n, \cdots$, or is empty.
Let $c_i$ be the number of a tile in the $i$-th cell in a state or $0$ if empty (the order in $i$ is arbitrary as long as neighborhood and directions are properly defined). 
Then, we denote state $s$ as a sequence of cells $s = [c_i]_{i\in[1,12]}$.  
Action space consists of four directions, $A(s) \subseteq \{\text{left, right, up, down}\}$, some of which can be invalid depending on the state. 
The transitions of \textit{2048} are well understood by \textit{afterstate} (Chapter 6.8 in the textbook by~\cite{SuttonBarto2018}). 
In \textit{2048}, after taking action $a_t$ at $s_t$, the state deterministically transits to afterstate $s'_t$. 
The transition causes all tiles to slide towards direction $a_t$, merging exactly two adjacent tiles with the same number along the direction if such tiles exist.  A merged tile has a new number that is the sum of two number tiles (i.e., double the old number) and gives an agent the reward of a new number.  
Then, the transition from afterstate $s'_t$ to the next state $s_{t+1}$ is stochastic; a new tile is placed at an empty cell with equal probability and its number is $2$ with 90\% or $4$ (with 10\%). 
The set of initial state $\mathcal{S}_0$ consists of a state with exactly two tiles placed by adopting the same procedure for the transition from afterstate to the next state.
If action $a_t$ does not involve change in any cell, i.e., afterstate $s'_t$ and state $s_t$ are equivalent, the action is invalid.  Consequently, a valid afterstate must have at least one empty cell as a result of sliding and/or merging. 
A game terminates when there is no valid action.

\begin{figure}[t]
  \subcaptionbox{an initial state\\\phantom{.}\hfill(age 4)\label{fig:initial}}[.24\linewidth]{\includegraphics[width=.9\linewidth]{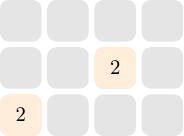}}
\subcaptionbox{afterstate after \textit{left}\\\phantom{.}\hfill(age 4)\label{fig:afterstate}}[.24\linewidth]{\includegraphics[width=.9\linewidth]{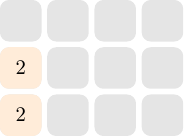}}
\subcaptionbox{one of next states\\\phantom{.}\hfill(age 8)\label{fig:nextstate}}[.24\linewidth]{\includegraphics[width=.9\linewidth]{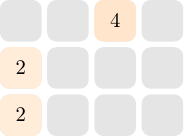}}
\subcaptionbox{a terminal state at best\\\phantom{.}\hfill(age 16380)\label{fig:terminalstate}}[.24\linewidth]{\includegraphics[width=.9\linewidth]{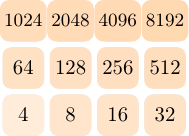}}
\caption{Example of states in \thisgame.  \textit{Age} is defined in Sect.~\ref{section:partition}}\label{fig:sample-state}
\end{figure}

Fig.~\ref{fig:sample-state} illustrates an example of initial state~(\ref{fig:initial}), and its afterstate after \textit{left} action~(\ref{fig:afterstate}).
After the transition, tiles were left aligned while there are no merged tiles in this example. 
Fig.~(\ref{fig:nextstate}) shows its possible next state by spawning a new tile.
The rightmost one~(\ref{fig:terminalstate}) is one of the best reachable terminal state.  There are no valid action because there are no empty cells or tiles can be merged in any direction. 

The goal of an agent is to maximize cumulative reward $\mathbb{E_\pi} \left[\sum_{t\ge 0} r_t \right]$ where the expectation covers the stochastic transition of MDP and the agent policy $\pi(a\mid s)$, which is the probability distribution (including a deterministic one) over actions conditioned on state $s$.\footnote{We treat discount factor $\gamma=1$ to be consistent with this game.}  
To strongly solve the game, we want an optimal policy $\pi^*(a\mid s)$ and optimal state value function $v^*(s_t) = \mathbb{E_{\pi^*}} \left[\sum_{t'\ge t} r_{t'} \right]$ presenting total rewards received in the future when the agent visit $s_t$ at time $t$.
Similarly, we define the optimal afterstate value function $v^*(s'_t)$.
Because the transition to afterstate is deterministic, denoting with $T$ as $s'_t = T(s_t, a_t)$, an optimal policy is defined to give probability one for action(s) with maximum $\max_{a\in A(s)}v^*(T(s, a))$ (ties can be distributed arbitrary) and zero for others, for each state $s$.\footnote{There can be multiple optimal policies though all of them share the same optimal value function.}  
For simplicity, we say the (after)state value of state $s$ for $v^*(s)$ if it is clear in the context that the value is for an optimal policy not subotpimal ones.  In short, we compute all afterstate values in \thisgame{} and store them in a retrievable manner so that one can easily identify an optimal action at runtime when visiting state $s$. 

\subsection{Related work}

The solutions of games have long been intensively studied as summarized by~\cite{HerikUiterwijkRijswijck2002}.  
A game is called \textit{strongly solved} when the values of all reachable states are identified and \textit{weakly solved} when the value of the initial state is identified, as in checkers by~\cite{SchaefferBurchBjornssonKishimotoMullerLakeLuSutphen2007} or in Othello by \cite{takizawa23:_othel_solved}.  
Many interesting games remain unsolved, and the number of valid positions can be very difficult to identify or even estimate, as in e.g., Go by~\cite{10.1007/978-3-319-50935-8_17} or chess\footnote{\url{https://github.com/tromp/ChessPositionRanking}}. 

In imperfect-information games,
different from perfect-information games or puzzles,
the goal is to obtain a set of strategies (policies) near Nash equilibrium. 
Although totally different techniques are needed for this purpose,
there is a similar challenge in how to manage huge state space. 
In methods based on counterfactual regret minimization (see e.g., \cite{MoravcikSchmidBurchLisyMorrillBardDavisWaughJohansonBowling2017,BrownSandholm2017b} for famous achievements),
\textit{counterfactual} values are maintained on each information state which is a set of worldstates composed considering hidden information, instead of state values in our case.

This study is in the line of adding a new single-player perfect-information game to strongly solved ones.  In general, stochastic games including \textit{2048} are more difficult than deterministic games because state values involve expectation over transition probabilities, disabling pruning techniques such as $\alpha\beta$ pruning effective in deterministic games. 

\section{Method}

We introduce the property of \textit{age} for states and afterstates, as a key idea.  Then, efficient enumeration of valid states leveraging their age and compact representation of results are described. 

\begin{figure}[t]
  \centering
  \includegraphics{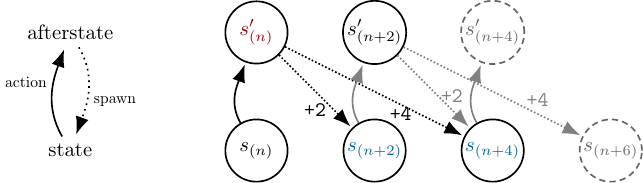}
  \caption{Transitions between state and afterstates by action and spawning of a new tile of $2$ or $4$, along with age $n$, i.e., each transition between consecutive states $s_t$ and $s_{t+1}$ increases age by $2$ or $4$.}
  \label{fig:transition}
\end{figure}

\subsection{Partition of states}\label{section:partition}

We define the \textit{age} of states and afterstates to partition them. 

\begin{definition}The \textit{Age} of state $s$ is the total of number tiles as
$\operatorname{age}(s) = \sum_{c_i\in s} c_i$, where we define state $s$ as a set of cells,
$s = [c_i]_{i\in[1,12]}$.  
The value of variable $c_i$ is $0$ if the corresponding cell is empty and the number of its tiles otherwise.
The age of afterstate $s'$ is defined similarly. 
\end{definition}

All valid states are partitioned by age, i.e., we can define a sequence of subsets of states, such that (1) all elements of each subset have the same age, (2) the intersection of any two subsets is empty, and (3) the union of all subsets includes all valid states. 
Similarly, we can partition all valid afterstates by age. 

Fig.~\ref{fig:transition} illustrates changes of age along with transition between a state and afterstate.  
Let $s_{(n)}$ and $s'_{(n)}$ be a state and after state with age $n$, respectively.  
Whenever a valid action leads $s_{(n)}$ to $s'_{(m)}$, we can confirm that their age is the same, i.e, $n=m$. 
It is straightforward if there are no merging tiles and from the fact that the number of new tiles is equivalent to the sum of the tiles merged otherwise. 
For transition from an afterstate to a state, age is increased by a new tile being added (noted by spawn in the figure). 
Note that although age $n$ monotonically increases along with time step $t$ as $n \ge 2t$, they do not form one-to-one mapping due to randomness in the number of tiles newly spawned.

Now it is clear that
\begin{enumerate}
\item a set of valid afterstates with age $n$, denoted by $\mathcal{S}'_n$, is identified by a set of valid states with age $n$, denoted by $\mathcal{S}_n$, and
\item a set of valid states with age $n$, $\mathcal{S}_n$, is identified by set of valid afterstates with two previous ages, $\mathcal{S}'_{n-2}$ and $\mathcal{S}'_{n-4}$.  
\end{enumerate}

\begin{figure}[t]
  \centering
  \includegraphics{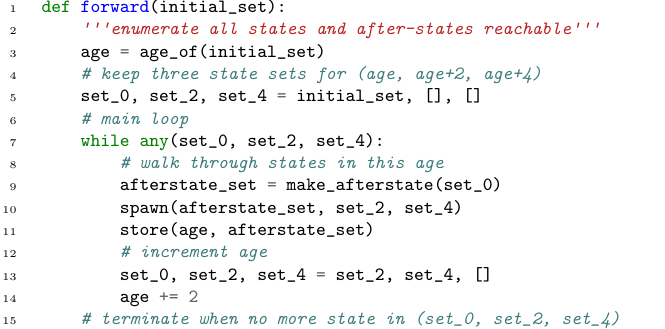}
  \caption{Pseudocode of \texttt{forward} procedure (borrowing Python syntax)}
  \label{fig:codeforward}
\end{figure}

Therefore, we can enumerate all valid states and afterstates by going through age $n$ in increasing order wihle keeping only three subsets of states in memory. 
Fig.~\ref{fig:codeforward} lists pseudocode (borrowing Python's syntax) of forward procedure for enumeration of (after)states, where the name is for increasing age. 
In the main \texttt{while} loop starting at line 7, we keep three state sets: \texttt{set\_0} for age, \texttt{set\_2} for age$+2$, and \texttt{set\_4} for age$+4$.
At the beginning of each loop \texttt{set\_0} is completed, but \texttt{set\_2} is partially enumerated, and \texttt{set\_4} is empty. 
This is because the age of a state after spawn depends on of tile number $2$ or $4$. 
At line 10, procedure \texttt{spawn} iterates over each empty cell in each afterstate with the current age and adds a state to \texttt{set\_2} by placing tile $2$ and similarly to \texttt{set\_4} by placing tile $4$.  More details of the spawn procedure will be discussed in Appendix. 
As listed in line 11, we suggest storing afterstates (without storing states) to save disk space. 

\begin{figure}[t]
  \centering
  \includegraphics{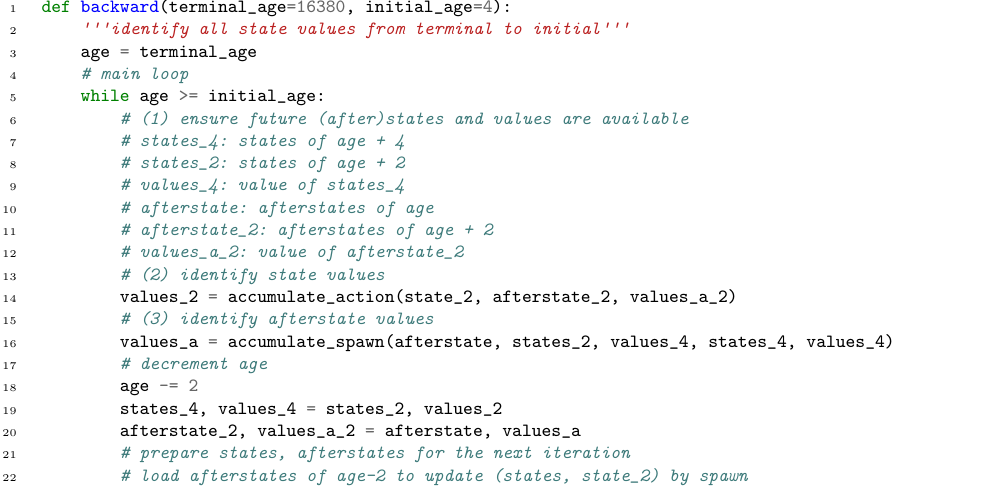}
  \caption{Pseudocode of \texttt{backward} procedure (borrowing Python syntax)}
  \label{fig:codebackward}
\end{figure}

To identify the value of each state or afterstate, we go backward in decreasing order of age. 
For this backward path, we observe
\begin{enumerate}
\item the value of each state with age $n$ is defined as the maximum value of corresponding afterstates, in $\mathcal{S}'_n$, and
\item the value of each afterstate with age $n$ is defined as the expectation of all corresponding states in $\mathcal{S}_{n+2}$ and $\mathcal{S}_{n+4}$. 
\end{enumerate}
Fig~\ref{fig:codebackward} illustrates pseudocode of the backward path.
At line 14, procedure \texttt{accumulate\_action} identifies values of states with age$+2$ by using that of afterstates with age$+2$.  
Then, at line 16, procedure \texttt{accumulate\_spawn} identify values of afterstates with age by using those of states with age$+2$ and age$+4$.  
The actual implementation of the omitted part is straightforward, but has a slightly tedious part that involves recovering a set of states from stored afterstates.

\subsection{Compact representation}

We need two efficient representations of a state or afterstate: for keeping unique IDs as a set in memory, and the other for storing in disk keeping look-up efficiency as high as possible. 

\subsubsection{ID for in-memory computation}\label{section:inmemory}
For in-memory computation, we adopted an unsigned 64-bit integer as (after)state identity, ID.
As shown in Fig.~\ref{fig:terminalstate}, the maximum number appearing on any tile is $2^{13}$.
Therefore, each cell $c_i$ is presented in a 4-bit integer $v_{c,i} \in \{0,1,\cdots,13\}$, 0 for empty cell $c_i=0$ or logarithm of its tile number $\log_2 c_i$ otherwise. 
Aggregating 12 cells in a state, we have a 48-bit unsigned integer. 
Because the number of states for each age is manageable, less than 300 million as shown in Fig.~\ref{fig:counts}, we did not adopt further compressionc during in-memory computation. 

With (after)state IDs, the forward and backward path require basically two functionalities; transition between (after)states and removing duplicates.  The former is straightforward as one can easily recover the status of each cell from (after)state ID.  
Regarding the latter, there are multiple sources of duplicates; different states may lead to the same afterstate after sliding and merging, and different afterstates may result in the same state by adding tile $2$ and $4$.  Also, 
equivalent states with symmetry (horizontal, vertical, and rotation of 180 degrees) should be unified. 
Moreover, parallel computation in each age may matter (implementation details will be discussed in Appendix). 
In our experience, the following straightforward way was most efficient; keep a sequence of 64-bit integers, generate new sequences ignoring duplicates, and then merge them after sorting of each vector (provided as \texttt{std::vector} and \texttt{std::set\_union} in C++\footnote{\url{https://en.cppreference.com/w/cpp/algorithm/set_union.html}}). 

\subsubsection{Compact representation in storage}
\begin{figure}[t]
  \centering
  \includegraphics{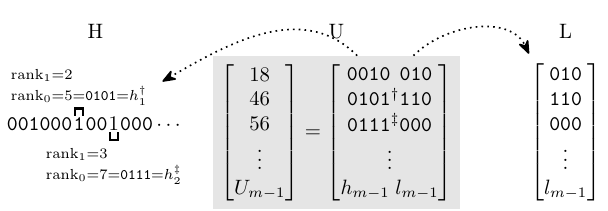}
  \caption{An illustrative example; set $U=(18, 46, 56, \ldots)$ whose values are less than $2^n\; (n=7)$ are presented as a pair of higher part $H$ with $q=4$ bits and lower $L$ with $n-q=3$ bits.  Binary digits are written in typewriter font.}
  \label{fig:sampleH}
\end{figure}

After enumeration of all (after)state IDs, we want to keep them in storage for the backward path as well as for further applications.  
A straightforward way is to save IDs as they are.  As the numbers of states and afterstates were about $10^{12}$ and $7\cdot 10^{11}$, they need $6.9$~TiB and $4.4$~TiB in total, respectively.\footnote{We assume 48-bit for each ID in estimation here.} 
While these sizes might be acceptable for current storages, they should preferably be reduced for further applications, e.g., evaluation of gameplay.  
Here, we introduce a classical technique about compact representation of a very sparse bit-vector supported by Elias-Fano codes. 
Here, we briefly introduce core techniques.  
Readers can safely jump to the Results section because the results did not depend on the techniques described here except for storage efficiency, where all afterstate IDs were stored in about $1.4$~TiB maintaining decent look-up efficiency.  
For further details of compact data structures, readers are referred to Chapter 4.4 of the textbook by~\cite{navarro16:_compac_data_struc_pract_approac}.

There are several equivalent understandings of the challenge on what to store; a sparse set of unsigned integers, a sequence of unsigned integers in strictly increasing order, a (conceptual) vary sparse bitset where the $n$-th bit is 1 if and only if number $n$ is included in the set in the previous problem.  We basically adopt the second interpretation.
Let $U$ be a sorted sequence of $m$ unique IDs that we want to store, $U_i$ be the $i$-th smallest element in $U$, and $n$ be a minimum integer satisfying $0 \le U_i < 2^n$ for all $i \in [0,m-1]$ (note that we intentionally adopt the $0$-index here).  Therefore, each $U_i$ is in $n$ bit length, resulting in $nm$ bits in total in a naive representation.
The core idea is to separate $n$ bits into a higher part $h_i$ with $q$ bits and a lower one $l_i$ with $(n-q)$ bits, such that $U_i = h_i \cdot 2^{n-q} + l_i$, where width $q$ is carefully chosen as $\lfloor \log_2 m \rfloor$.  Hereafter, we omit floor symbols $\lfloor \rfloor$ for simplicity in notation.
Then, the lower $l_i$s are concatenated into single bit-vector $L$ of length $(n-q)m$ bits and stored as is. 
The higher bits $h_i$s are embedded into a well crafted bit-vector $H$. 
Important properties of $H$ are:
\begin{itemize}
\item $H$ is bit-vector of $(m + q)$ bit-length, containing $m$ $1$s and $q$ $0$s,
\item bit 1 in $H$ is set for each position $k + h_k$ for $0 \le k < m$,\footnote{The positions are distinct because $u_k$ and consequently $h_k$ are sorted in increasing and non-decreasing order, respectively.} and consequently,
\item for any $j$ and $k$ such that the $j$-th bit in $H$ is the $k$-th 1 from the beginning, i.e., $H[j]=1$ and $\text{rank}_1(H, j)=k$, then, value $h_k$ is given by $\text{rank}_0(H, j)$, 
\end{itemize}
where functions $\text{rank}_0(H, j)$ and $\text{rank}_1(H, j)$ are defined as the number of $0$s and $1$s in the first $j$ bits in bit-vector $H$, respectively. 
In other words, one can visit all $h_k$s by iterating each bit in $H$, from the beginning to the end, with counting of $0$s and $1$s. 
Note that, although items in the original set $U$ are unique, their higher bits may have the same value and these properties are consistent with such cases.
The reduction in size comes from data structure $H$ because the lower part $L$ is stored as it is.
While naive storing of the higher part requires $qm$ bits ($m$ items and $q$ bits each), 
we stored $H$ in $m + q$ bits.  With $q = \log_2 m$, the difference in size is $qm - (m + q) = (m-1)\log_2 m - m > 0 \; (m\ge 3)$.  
Note that the sparseness, $m \ll 2^n$, is crucial as $m \approx 2^n$ indicates $q \approx n$ where all bits of items will be stored in $H$ in a redundant manner.
Fig.~\ref{fig:sampleH} illustrates an example of set $U=\{18, 46, 56, \cdots\}$ where each element in $U$ is presented in $n=7$ binary digits.   Assuming that $q=4$, we separate each item into its higher $4$ bits and the other $3$ bits.   The latter parts are concatenated into bit vector $L$.   The higher bits are stored in $H$ where each bit with value 1 has one-to-one mapping of each item in $U$, as shown in illustration of items $h_1=\texttt{0101}^\dagger$ and $h_2=\texttt{0111}^\ddagger$.

We adopted three additional tweaks to have smaller bit lengths $n$ for IDs, while the number of IDs, $m$, is constant. 
To do so, we want to bound the maximum ID as small as possible. 
Simply choosing the smallest one among symmetries works.  Also, we can change the base depending on the largest tile number. 
So far, we encode state ID as $\sum_{i\in [1, 12]} v_{c,i}\cdot 16^{i-1}$ giving each cell 4-bit width.  However, given that the maximum value of $v_{c,i}$ is 13, it is safe to encode ID with base $b=14$ as $\sum_{i\in [1, 12]}v_{c,i}\cdot b^{i-1}.$  
Base $b$ can be decreased more in accordance with the maximum value for each age. 
The last tweak is to drop information about cell $c_1$ to formulate compressed ID as $\sum_{i\in [\textbf{2}, 12]}v_{c,i}\cdot b^{i-2}.$  
Recall that we partitioned (after)states by their age, so, the sum $\sum_{i\in [1, 12]}c_i$ is constrained for each partition.  Therefore, we can safely forget exactly one cell and later recover all information. 

\begin{table}[t]
  \centering
  \caption{Example of stored afterstates: \#IDs of IDs are stored in $H+L$ bytes much smaller than in-memory size, which is 8byte/item.}\label{tab:astate8000}
  \begin{tabular}{rrrrrrrrr}\toprule
    \multicolumn{1}{c}{age}  & \multicolumn{1}{c}{\#IDs ($m$)} & \multicolumn{1}{c}{base ($b$)} & \multicolumn{1}{c}{hi ($q$)} & \multicolumn{1}{c}{low} & \multicolumn{1}{c}{$H$} & \multicolumn{1}{c}{$L$} & \multicolumn{1}{c}{in memory size} & \multicolumn{1}{c}{ratio}\\
    & & & (bits) & (bits) & (bytes) & (bytes) & (bytes) &  (\%)\\\midrule
    2000 & $75,344,033$ & 12 & 26 & 14 & $17,952,888$ & $114,771,256$ & $602,752,264$ & 22.0 \\ 8000 & $78,982,989$ & 14 & 26 & 14 & $17,384,728$ & $125,945,520$ & $631,863,912$ & 22.7 \\ \bottomrule
  \end{tabular}
\end{table}

Table~\ref{tab:astate8000} lists two examples of storage size for age 2,000 and 8,000.
Column \#IDs $m$ is for the number of afterstates in each age, already shown in Fig.~\ref{fig:counts}. 
For the width of the higher part, we used $26$ as the floor of $\log_2 m$, though the results are similar if the ceiling is adopted instead of the floor.  The actual storage size of $H$ and $L$ are slightly larger than the theoretical value due to 64-bit boundary with a sentinel.  Compared to bytes in memory used in Sect.~\ref{section:inmemory}, the storage size of the sum of $H$ and $L$ is about 22\%, and we believe that it is well compressed. 

Overall, one can look up an afterstate by the following procedure:
\begin{enumerate}
\item identify its age as the sum of all tile numbers, 
\item encode ID as 48-bit unsigned integer, normalize it as the minimum ID in the equivalent symmetries,
\item convert it to compressed ID $u$ by dropping the least significant digit, and rebasing with base $[12,14]$ depending on the age
\item separate $u$ as pair of higher and lower bits $(u_h, u_l)$, such that $u=u_h \cdot 2^{(n-q)} + u_l$ where width $(n-q)$ depends on the age
\item lookup storage $H$ of the age to find the minimum index $i_h$ satisfying $\text{rank}_0(H, i_h-1)=u_h$.  If the index is more than its length then the result is not found.  Otherwise,
\item let $i_l=\text{rank}_1(H, i_h)$ and lookup storage $L$ of the age to see whether the $i_l$-th value equals $u_l$.  If true, return success.  Otherwise, check if $(i_h+1)$-th bit is 1 in $H$.  If so, because it means multiple items in $U$ share the same upper bits $u_h$, continue with incremented $i_h$ and $i_l$.
\end{enumerate}

For (after)state values, we simply store values following the order of afterstates, at each age.   
Therefore, it is straightforward to retrieve the (after)state value by using its age and rank $i_l$.
It is known that by adding appropriate indices, rank and select operations on $H$ becomes more efficient.  
However, we simply adopted a linear search with 64-bit pop-count operation equipped with modern CPUs because it is efficient enough compared to the cost of disk read for $L$ or values. 

\subsubsection{Representation for optimal playing}\label{sect:burr}

For optimal playing, one can compress more to have an optimal move for each state less than $275$~GiB total. 
Recent advances enable us to represent a set of (after)states with $1.5$ bits per item and a state-value mapping with $1.01\cdot r$ bits per item where $r$ is bit length of the value for each state (see, e.g., survey by \cite{lehmann2025modernminimalperfecthashing}), if one can assure a valid (after)state is given with each query.  This is much more efficient than using more than ten bits as in our empirical data listed in Table~\ref{tab:astate8000}.  
For optimal playing, we want to lookup an optimal action for a state, where 
$r=2$ bits are enough for at most four valid actions. 
In fact, in our experiences, the BuRR algorithm, by ~\cite{dillinger_et_al:LIPIcs.SEA.2022.4} with publicly available implementation,\footnote{\url{https://github.com/lorenzhs/BuRR}, \url{https://github.com/ByteHamster/SimpleRibbon}} successfully stored all optimal moves within about $1.9$~bits per state.  That is, the storage for states themselves is almost negligible while values available are kept for any valid state in a query. 

A drawback of these techniques is that one cannot recover or enumerate (after)states from the database itself and needs to an (after)state give with each query.  This prevents some applications requiring access to the entire (after)state space, e.g., study on exploring starts in reinforcement learning.  
Note that our backward procedure also requires entire afterstates at each age.  Therefore, it is suggested to keep the original until its completion.

\section{Results}

\begin{figure}[t]
  \includegraphics[width=\linewidth]{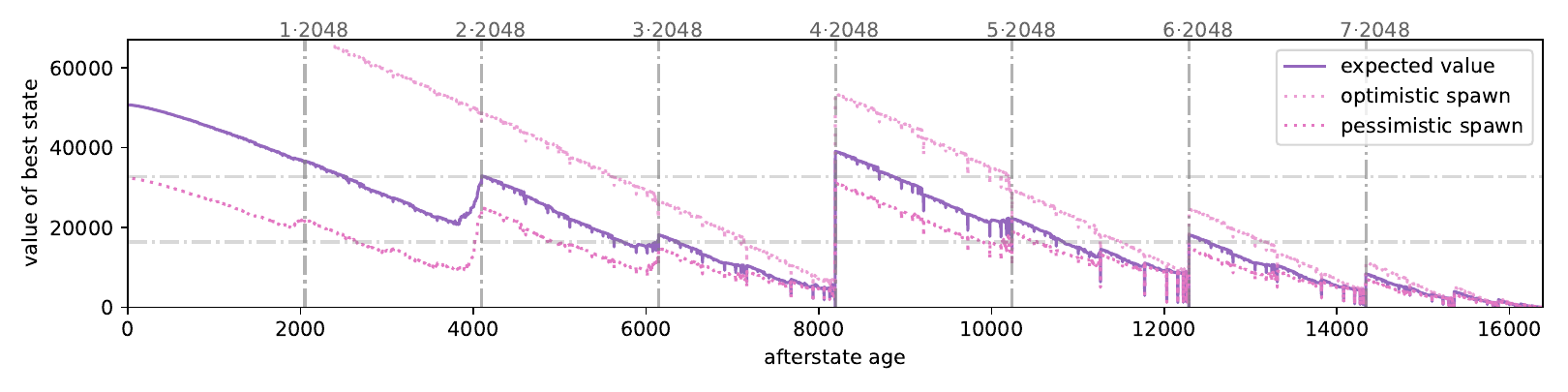}
  \caption{Value (expected returns in future) of best afterstates at each age (purple).  Peaks at age near multiples of 2048 indicate difficulty of making a tile of larger numbers.  Two dotted lines (pink) are for optimistic and pessimistic variation assuming each tile spawns at the most and least preferable cell in the future, respectively.  The former starts $82705.6$ at age 4 though $y$-axis is arranged to better focus on primary values.}
  \label{fig:values}
\end{figure}

\begin{figure}[t]
  \includegraphics[width=\linewidth]{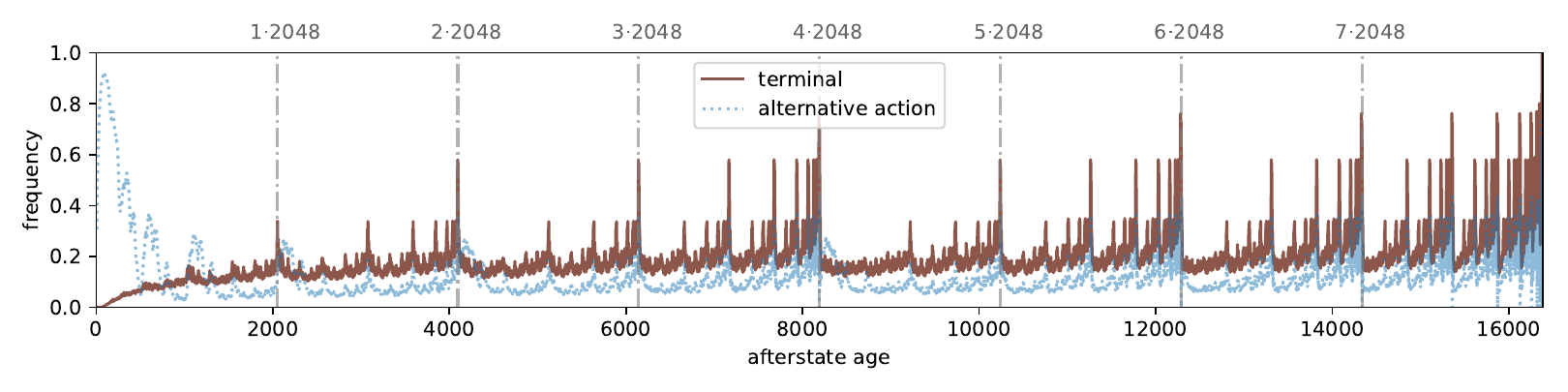}
  \caption{Relative frequency of afterstates being terminal at each age (brown).  The ratio is around $0.2$ for most ages, but there are occasional peaks, meaning that one needs careful arrangement of tiles to go beyond that age.  Dotted line (blue) shows the frequency of states having an effective alternative action (see main text for details). }
  \label{fig:zero-values}
\end{figure}

Fig.~\ref{fig:counts} already shows the number of reachable states and afterstates (the vertical axis) for each age (the horizontal axis), which are identified by the forward path (listed in Fig.~\ref{fig:codeforward}). 
Multiple valleys were observed at ages multiple of 2048, shown by vertical dashed lines.
This is consistent with common sense: making tiles of 2048 or more is difficult because it requires careful arrangement of $n$ tiles to make a tile of $2^n$ while the number of available cells is limited.  
Related to this topic, there are no valid afterstates at age $8190, 12286, 14334, 15358, 15870, 16126, 16254, 16318, 16350, 16366, 16374,$ or $16378$.  
This is because at least 12 tiles are needed to make the sum of any of these numbers while a valid afterstate can have at most 11 tiles.
We further confirm the consistency in terms of state values.

The backward path (listed in Fig.~\ref{fig:codebackward}) identified the value of each (after)state (that is, expected returns in the future), starting at the afterstate.  
Fig.~\ref{fig:values} shows the values of the best afterstates at each age the in primary line (purple).  
The value takes about $50724.26$ at age 4, initial states, and roughly decreases as age increases with occasional \textit{jumps} at ages multiple of 2048, indicating that one needs some luck to make a large number of tiles. 
It is natural that the difficulty of obtaining more rewards avoiding termination increases when a game proceeds in general and just before making a large number of tiles. 
Recall that state value does not include past rewards obtained before reaching the state.  
The other two dotted lines (pink) in Fig.~\ref{fig:values} show optimistic and pessimistic variants of values assuming that each tile spawns the most and least preferable cell, respectively.   After its number is sampled with the original probability, i.e., $2$ with 90\% or $4$ with 10\%, the cell with the maximum and minimum values (of that variant) is chosen.  
The gaps between primary (purple) and the dotted (pink) lines are small near terminal or ages before making a large tile.  
Please note that there are other interesting analyses with different definitions because this is our interpretation of an optimistic or pessimistic situation and an example demonstrating that our method consistently works with slight changes in an environment. 
The difficulty of making a large tile can also be illustrated by relative frequency of afterstates being terminal, i.e., there are no more valid actions after spawning a new tile, as shown by the primary line (brown) in Fig.~\ref{fig:zero-values} for each age. 

For initial states, there are three cases, age 4 (two tiles of $2$), age 6 (one tile $2$ and one tile $4$), and age 8 (two tiles of $4$).   We found that all initial placement and actions are equivalent for each of these cases. 
For the first case, all afterstates with two $2$ tiles have value $50724.26$, and all afterstates with one $4$ tile (by merging two $2$ tiles) have value $50720.26$, where by adding an immediate reward of $4$ by merging tiles we have the same expected score.
For the second case, all initial placement and actions are equivalent with value $50720.62$.  
This means that there is an about $3.64$-point disadvantage to picking a tile $4$ in an initial position. 
For the last case, the value is $50716.99$ without merges, or $50708.99$ after obtaining reward 8 by merging two $4$ tiles, for all states.  In this case, disadvantage is roughly $7$ points.
Therefore, in summary, having tile $4$ at an initial state has a small disadvantage in the expected score. 

\begin{figure}[t]
  \includegraphics[width=\linewidth]{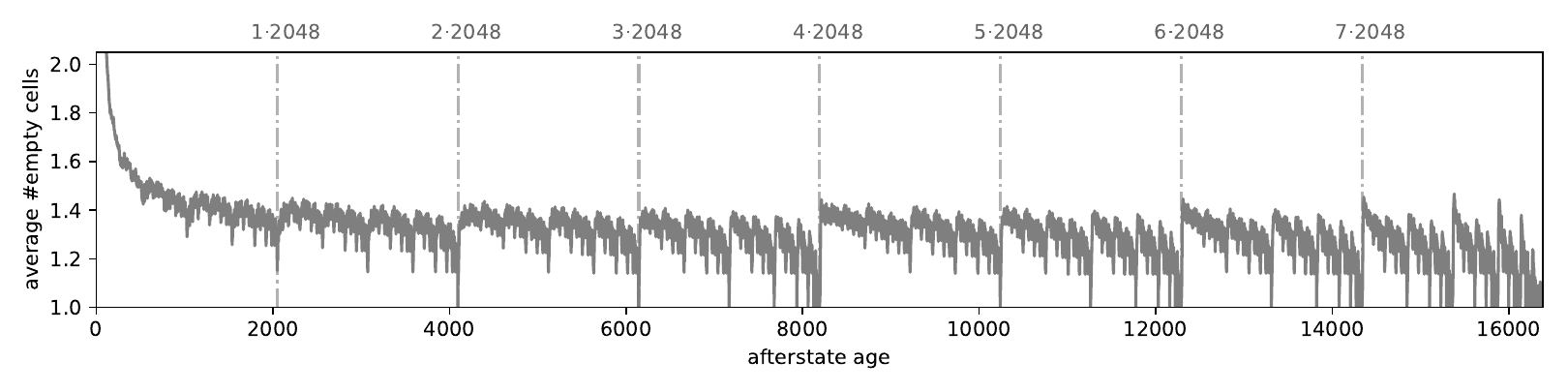}
  \caption{Average number of empty cells at each age.  Note that a valid afterstate has at least one empty cell.}
  \label{fig:empty-cells}
\end{figure}

Fig.~\ref{fig:empty-cells} shows the average number of empty cells at each age.
Interestingly, the values are rather stable under $1.5$ except for the very beginning of a game, e.g., about $2.01$ at age 120. 
Note that there must be at least one empty cell according to the game rules.

\begin{figure}[t]
  \includegraphics[width=\linewidth]{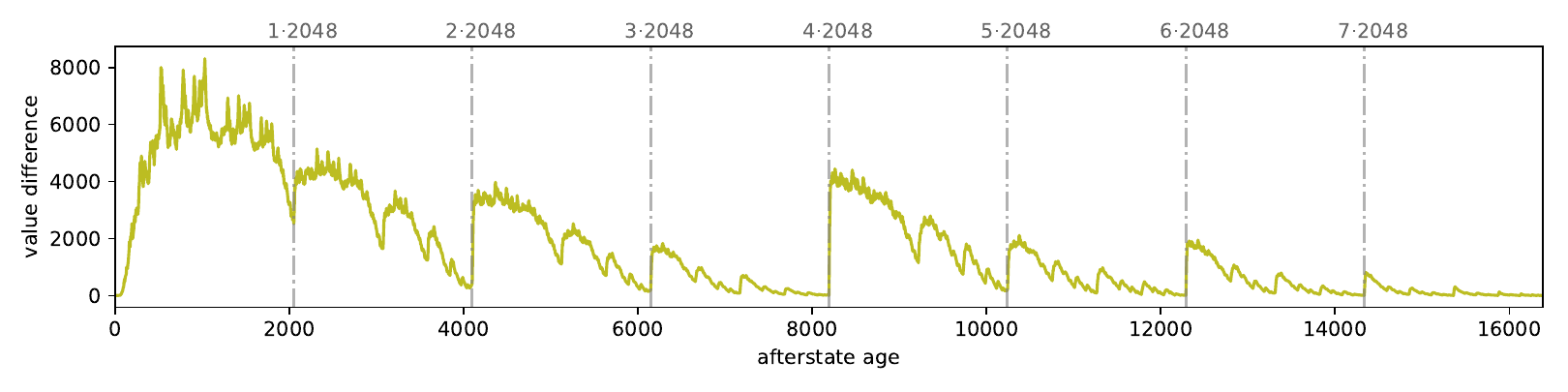}
  \caption{Mean difference in values between optimal action and suboptimal one.}
  \label{fig:qdiff}
\end{figure}

As a potential application of the obtained values, 
we additionally investigated the difference between optimal and sub-optimal actions.
Fig.~\ref{fig:qdiff} shows the mean difference for each age. 
There are clear peaks near age 2048 and 4096, indicating the importance of the first several actions after obtaining a tile with large numbers.  
It would support empirical understanding of why separation of evaluation functions for stages worked well in past studies,~\cite{WuYehLiangChangChiang2014,Jaskowski2017}. 
Relatively large differences in general indicate that an agent need to follow the best action in most cases to achieve a good score. 
This observation is consistent with the frequency of afterstates where the difference is less than $1.0$ shown by the dotted line (blue) in Fig.~\ref{fig:zero-values}.

\section{Conclusion}

This paper presented the solution for \thisgame, a $4\times 3$ variant of \textit{2048}, which is a stochastic single-player game. 
In this variant, the expected score achieved by an optimal strategy isabout $50724.26$ for the most common initial states with two tiles of number 2 and few points lower for other initial states.  
The number of afterstates was $739,648,886,170$, and all IDs were stored in about $1.4$ TiB. 
The key technique enabling \thisgame{} to be solved is to leverage a game-specific property of \textit{age}, the sum of the numbers on a board.  Age is invariant between a state and its successive afterstate after any valid action and is increased two or four by stochastic response from the environment. 
State (and also afterstate) spaces are clearly partitioned by age, enabling a huge reduction in memory usage. 
We also adopted a compact representation of state ID set on the basis of Elias-Fano code, which will hopefully work well in holding state IDs in other games where ranking/unranking functions are not known. 
By using these techniques, the optimal value function of \thisgame{} is now available for researchers with decent computational resources, about $4$ TiB ssds and few days of computation with a modern CPU. 
Also, data specialized for optimal playing only requires about $300$~GiB.  
The authors believe that the dataset in \thisgame{} as well as core ideas will serve as a baseline of further studies.

\configurebibstyle{}
\bibliography{ace,learning}

\begin{appendix}
\section{Implementation details}

\begin{figure}[t]
  \centering
  \includegraphics{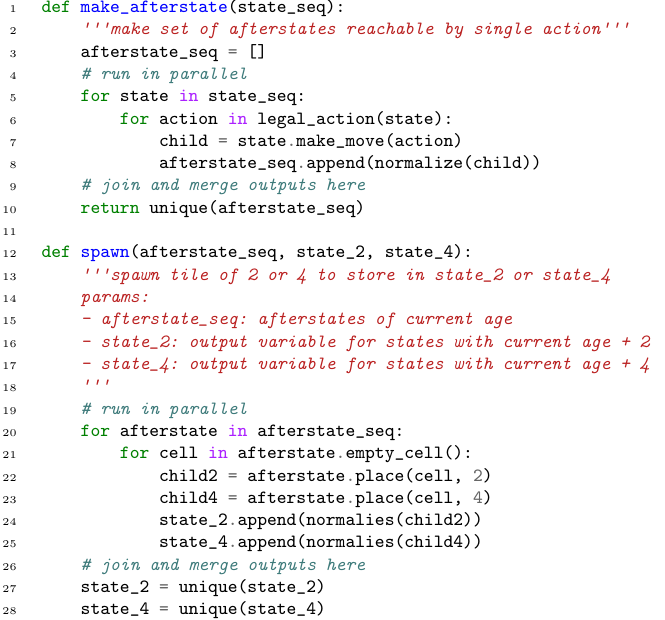}
  \caption{Miscellaneous functions used in \texttt{forward} (borrowing Python syntax)}
  \label{fig:codemisc}
\end{figure}

\begin{figure}[t]
  \centering
  \includegraphics{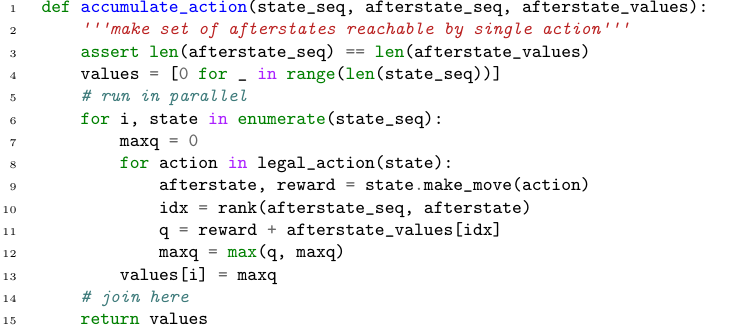}
  \caption{Miscellaneous functions used in \texttt{backward} (borrowing Python syntax)}
  \label{fig:codemiscback}
\end{figure}

In the main experiments, the forward and backward computation require about 25 and 40 hours, respectively,
implemented in C++ ran on a Linux server with AMD Ryzen 9 7950X 16-Core Processor. 
The memory usage was about $32$~GiB for full (16 cores) parallelization. 
Binaries are stored via a \texttt{mmap} system call on Linux, so, they are machine dependent.
For detecting bit errors, we used PicoSHA2 to obtain hash digests with SHA256 for state IDs.
Also, the consistency of the results was confirmed by three independent runs of the whole forward and backward procedures.

Here are some details on our forward and backward functions listed in Figs~\ref{fig:codeforward} and \ref{fig:codebackward} in the maintext. 
The first function in Fig~\ref{fig:codemisc} illustrates how \texttt{make\_state\_after\_action} is implemented, borrowing Python syntax for simplicity. 
The function produces a sequence of afterstate IDs reachable from state IDs given in \texttt{state\_seq}.  
The outer \texttt{for} loop in line 5 may run in parallel with changing output variable \texttt{after\_state\_seq} thread-local.  
The outputs from all threads are then merged into a single sequence and sorted while removing duplicates.  
To remove duplicates, IDs should be normalized to resolve symmetries in advance.
Similarly, the second function \texttt{spawn} takes a sequence of afterstate IDs and outputs successor state IDs.  Here, two sequences are separately maintained in the output for spawn of tile $2$ and tile $4$. 
The outer \texttt{for} loop in line 20 may run in parallel, similarly. 

In the backward path, two functions (\texttt{accumulate\_action} and \texttt{accumulate\_spawn}) compute values of states and afterstates, respectively, one half-step from an older age to a younger one.  
Fig.~\ref{fig:codemiscback} gives pseudocode for \texttt{accumulate\_action} that identifies all values of states taking input of successor afterstates and their values as sequences ordered consistently.
For each state, action value \texttt{q} is identified as the sum of immediate reward \texttt{reward} by merging tiles (if any) and afterstate's value.  
Function \texttt{rank} of \texttt{after\_state} in the inner loop is given by \texttt{std::lower\_bound} leveraging increasing order of \texttt{afterstate\_seq}.  
The outer \texttt{for} loop at line 5 may run in parallel similarly as functions above.
Function \texttt{accumulate\_spawn} is similarly defined except for taking the expectation instead of the maximum. 

The disk space required for afterstates was about $1.4$ TiB and depends on conditions for values.
Given that the game involves probability $0.1$ at each step and may reach more than 8000 steps, it is reasonable to introduce approximation by floating point numbers.  Therefore, we used the standard 64bits for in-memory computation. 
In storing those values, we simply kept precision of $2^{-16}$ by scaling $2^{16}$ before storing as integer. 
Considering the integer part of all values presented in 15 bits or less for most ages as shown in Fig.~\ref{fig:values}, 
this simple approach required about $2.56$~TiB total. 
The scaling factor can be adjusted to balance between precision and space.  For example, with precision $2^{-8}$,
storage is about $1.89$~TiB. 
Note that one needs only $300$~GiB when specialized for optimal playing, as introduced in Sect.~\ref{sect:burr}. 
The authors' implementation is available at \url{https://github.com/tkaneko/db2048-4x3}. 

\end{appendix}
\end{document}